\documentclass[sigconf]{acmart}
\AtBeginDocument{%
  \providecommand\BibTeX{{%
    \normalfont B\kern-0.5em{\scshape i\kern-0.25em b}\kern-0.8em\TeX}}}

\setcopyright{acmcopyright}
\copyrightyear{2023}
\acmYear{2023}
\acmDOI{XXXXXXX.XXXXXXX}

\acmConference[Conference acronym 'XX]{Make sure to enter the correct
  conference title from your rights confirmation email}{June 03--05,
  2023}{Woodstock, NY}
\acmPrice{15.00}
\acmISBN{978-1-4503-XXXX-X/23/06}

\usepackage{multicol}
\usepackage{multirow}
\usepackage{bbding}
\usepackage{amsfonts} 
\usepackage{amsthm}
\usepackage{hyperref}

\begin{document}

\title{Intelligent Virtual Assistants with LLM-based Process Automation}

\author{Yanchu Guan$^{1,* }$, Dong Wang$^{1,*}$, Zhixuan Chu$^{1,*}$, Shiyu Wang$^{1,*}$}
\author{Feiyue Ni$^2$, Ruihua Song$^2$, Longfei Li$^1$, Jinjie Gu$^1$, Chenyi Zhuang$^{1,\dag}$}
\affiliation{%
  \institution{$^1$Ant Group}
  \country{}
  \institution{$^2$Renmin University of China}
  \country{}
}
\email{{yanchu.gyc, yishan.wd, chuzhixuan.czx, weiming.wsy, longyao.llf, jinjie.gujj, chenyi.zcy}@antgroup.com}
\email{nifeiyue@ruc.edu.cn,songruihua_bloon@outlook.com}

\renewcommand{\shortauthors}{Yanchu, Dong, Zhixuan, and Shiyu, et al.}

\begin{abstract}
  
While intelligent virtual assistants like Siri, Alexa, and Google Assistant have become ubiquitous in modern life, they still face limitations in their ability to follow multi-step instructions and accomplish complex goals articulated in natural language. However, recent breakthroughs in large language models (LLMs) show promise for overcoming existing barriers by enhancing natural language processing and reasoning capabilities. Though promising, applying LLMs to create more advanced virtual assistants still faces challenges like ensuring robust performance and handling variability in real-world user commands. This paper proposes a novel LLM-based virtual assistant that can automatically perform multi-step operations within mobile apps based on high-level user requests. The system represents an advance in assistants by providing an end-to-end solution for parsing instructions, reasoning about goals, and executing actions. LLM-based Process Automation (LLMPA) has modules for decomposing instructions, generating descriptions, detecting interface elements, predicting next actions, and error checking. Experiments demonstrate the system completing complex mobile operation tasks in Alipay based on natural language instructions. This showcases how large language models can enable automated assistants to accomplish real-world tasks. The main contributions are the novel LLMPA architecture optimized for app process automation, the methodology for applying LLMs to mobile apps, and demonstrations of multi-step task completion in a real-world environment. Notably, this work represents the first real-world deployment and extensive evaluation of a large language model-based virtual assistant in a widely used mobile application with an enormous user base numbering in the hundreds of millions.
\end{abstract}





% \received{20 February 2023}
% \received[revised]{12 March 2023}
% \received[accepted]{5 June 2023}

\maketitle

\def\thefootnote{$\ast$}\footnotetext{These authors contributed equally to this work.}

\section{Introduction}

In modern times, intelligent virtual assistants such as Siri, Alexa, and Google Assistant have become widespread and pervasive in people's daily lives. However, despite their prevalence, these artificial intelligence systems still face constraints in their capacity to carry out intricate multi-step procedures for their human users. Nevertheless, with the swift advancement and evolution of large language models (LLMs) \cite{brown2020language,openai2023gpt4,touvron2023llama,chu2023datacentric}, there is optimism that these LLMs may help conquer the existing limitations by comprehending natural language directions more profoundly, applying logic to identify objectives, and independently orchestrating sequences of activities. By enhancing causal reasoning abilities \cite{li2023machine}, large language models could enable virtual assistants to understand ambiguous instructions, break down complex goals into executable steps, and autonomously complete chained tasks to fulfill user requests \cite{dong2023towards}. The continued progress in large language model research shows promise for overcoming the boundaries of today's virtual assistants.

In this paper, we propose a novel intelligent virtual assistant system based on LLM-Based Process Automation (LLMPA), which can automatically perform operations within mobile applications based on high-level user requests. Unlike prevalent virtual assistants which rely heavily on invoking fixed programmatic functions, our proposed system directly emulates detailed human interactions for carrying out tasks. This human-centric approach grants greater adaptability to perform more unconstrained, multi-stage procedures based on natural language directions. While prevailing assistants like Siri and Alexa can only execute simplistic pre-defined behaviors through rudimentary API calls, mimicking human-like actions empowers our assistant to operate at a higher level of abstraction. By simulating granular operations like clicks, scrolls, and types, our agent can flexibly conduct these operations. This enables accomplishing substantially more complex goals involving free-form instructions, creative problem-solving, and generalized tasks beyond the rigid constraints of current assistants' underlying frameworks. 

The proposed LLM-Based Process Automation (LLMPA) is central to our approach, with modules for decomposing instructions, generating natural language descriptions, detecting interface elements, predicting next actions, and checking for errors. We demonstrate our system using the Alipay mobile payments app as a target environment. Users can simply describe a high-level task, such as ordering a coffee, and our system will automatically navigate the app, select items, enter information, and make payments as needed. This showcases how large language models can enable automated mobile assistants to carry out complex real-world tasks based on natural language instructions and environmental context.

The main contributions of our work are the novel LLMPA model architecture optimized for app automation, the methodology for applying LLM-based assistants to mobile apps, and demonstrations of multi-step task completion in a real-world environment. Notably, this work represents the first real-world deployment and extensive evaluation of a large language model-based virtual assistant in a widely used mobile application with an enormous user base numbering in the hundreds of millions. By successfully demonstrating complex multi-step task completion capabilities in the massively popular Alipay platform, our system marks a major milestone in translating large language model research from controlled experimental settings into large-scale practical applications with tremendous reach and impact. The expansive testing of our approach in a real production environment at this scale is unprecedented in the field of intelligent assistants, underscoring the significant advancements enabled by modern large language models toward building assistants that can reliably understand instructions, reason about goals, and accomplish procedural tasks to aid millions of end users.

\section{Related Work}
For a while now, artificial intelligence has aimed to develop agents that possess general intelligence and can perform cognitive tasks like humans \cite{wang2023enhancing,chu2023leveraging}. Ideally, these agents should be capable of communicating through natural language and solving any computer task that a human can. By automating repetitive tasks and offering support in intricate problem-solving, these virtual agents could significantly enhance productivity.

In recent years, there have been several advancements in the field of autonomous agents as virtual assistants. For example, Apple introduced Siri, a voice assistant that helps users automate various tasks through interactive agents. Amazon launched Alexa, a virtual assistant that serves as a home automation system to control multiple smart applications or devices and perform various tasks. Google also introduced Google Assistant, an automated agent for human-computer interaction, aimed at enabling users to communicate with devices seamlessly through natural language.

It is worth noting that the combination of Large Language Models (LLMs) and Autonomous Agents has become a popular trend, largely due to the recent success of large language models. By harnessing LLMs, we can effectively handle natural language inputs and perform logical reasoning to understand user intent, enabling us to act as virtual agents for various complex tasks. This trend has greatly encouraged the exploration of LLM-augmented Autonomous Agents (LAAs) in real-world applications. For example, MindACT \cite{deng2023mind2web} is a web agent that can follow language commands to perform complex tasks on any website. MetaGPT \cite{hong2023metagpt} is an innovative framework that incorporates efficient human workflows as meta-programming methods into LLM-based multi-agent collaboration. Flowris \cite{sun2023flowris} is a conversation agent focused on data analysis, which can collect and manage source data of conversation agents for data analysis and management purposes.

Nevertheless, it is imperative to acknowledge that prevailing approaches encounter numerous obstacles. The core requirement for autonomous agents is the ability to accurately understand user intent and automatically generate corresponding actions. Therefore, we present a novel approach that utilizes instruction chain technology to refine the LLM and engender manageable sequences of actions. By harnessing the formidable capabilities of LLM, we acquire profound insights into the contextual backdrop of users, thereby facilitating the prognostication of their forthcoming actions. Furthermore, cognizant of the inherent indeterminacy in generated actions, we have built a controllable calibration module to scrutinize the logical coherence of the action sequences.

\section{FRAMEWORK}
\begin{figure}[h!]
    \centering    \includegraphics[width=0.45\textwidth]{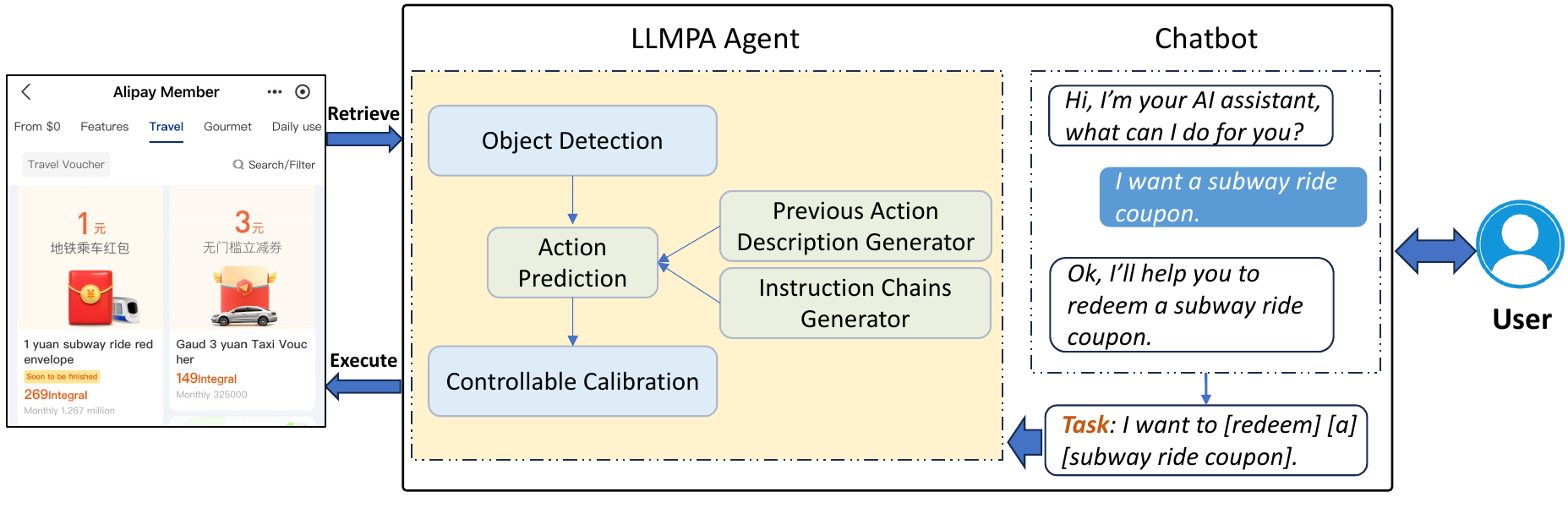}
    \caption{Architecture of Intelligent Virtual Assistants that includes LLMPA Agent and Chatbot.}
    \label{fig:system_overview}
\end{figure}

\begin{figure*}[h!]
    \centering    \includegraphics[width=0.95\textwidth]{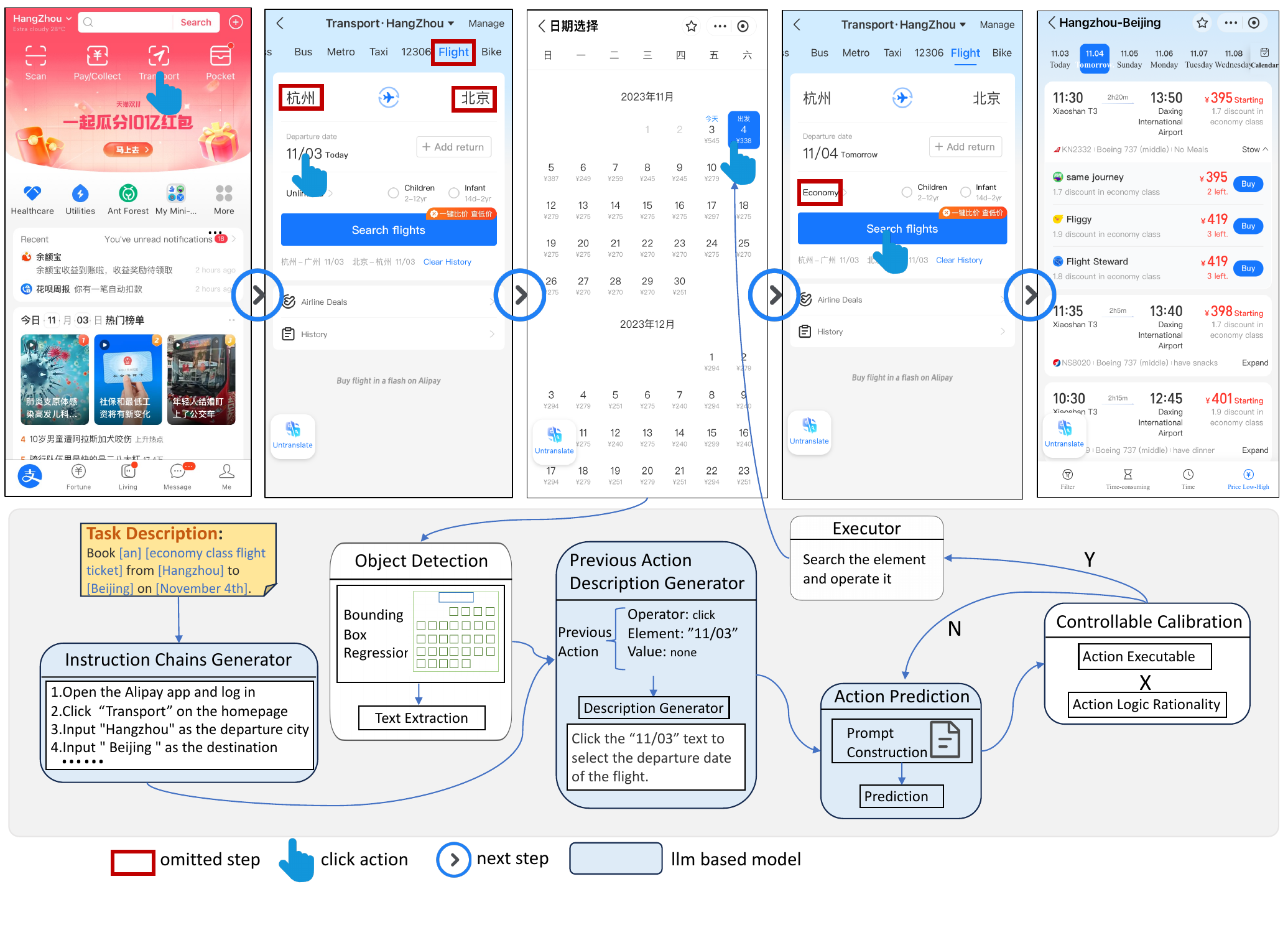}
    \caption{Pipeline of LLMPA Agent.}
    \label{fig:system_pipeline}
\end{figure*}

This section provides an overview of the proposed Intelligent Virtual Assistants. As shown in Figure 1, the User engages with the Chatbot, clearly outlining the objective. The LLMPA Agent collaborates with the app, aiding the user in accomplishing the operations.

The Chatbot of our proposal includes a multi-turn dialogue module and an intent extraction module. In simple terms, it is designed to comprehend user requirements and generate appropriate task descriptions. Currently, a lot of research~\cite{adamopoulou2020overview,zhang2020recent,li2020conversation,cahn2017chatbot} is being conducted on the topic, which is not the focus of our paper. 

The LLMPA Agent of our proposal is dedicated to understanding the task, deconstructing it, and then methodically executing it. This agent includes the following modules:
1) \textbf{Instruction Chains Generator.} This module decomposes the task and produces detailed step descriptions.
2) \textbf{Previous Action Description Generator.} Based on the prior action and the page content, this module produces an intelligible description~\cite{qin2021intelligible} of the action.
3) \textbf{Object Detection.} We introduce an object detection \cite{zou2023object} model for the section recognition of the page. The text within a section is classified into a group, which presents a clear hierarchical structure, contributing to enhanced comprehension of the context.
4) \textbf{Action Prediction.} Based on the outputs of preceding modules, we can construct a prompt \cite{white2023prompt,xue2023prompt,liu2023pre}, directly predicting the subsequent action.
5) \textbf{Controllable Calibration.} Many studies \cite{zhang2023siren,bang2023multitask,dhuliawala2023chain} have been discussing the hallucination phenomena in large language models(LLMs). To mitigate the impact of the hallucination issue, we designed a controllable calibration module. This module is utilized to scrutinize the predicted action, ensuring that the action is operable.

\section{Methodology}

In this section, we choose the example of ``booking a flight ticket on Alipay'' to illustrate how the LLMPA Agent works. And we will explain every module in detail, following the pipeline shown in Figure 2. Given the extensive length of the complete flight booking procedure, we display only a portion of the steps involved. Steps that have been omitted are conveniently marked with red boxes. Since the task unfolds in a loop with each step adhering to the same procedure, we elect to illustrate this process using the ``select flight date'' step as an example. Let’s start with some notations and definitions.

\subsection{Notations and Definitions}
In this section, we give the following definitions used in the paper.

Definition 1. \textbf{action.} 
Let $\mathcal{A}$ denote the action space. For any action $a \in \mathcal{A}$, we have $a=f(e, v)$. Here, $e$ represents the element, which can be any text on the page. $v$ stands for value and $f$ means the function, where $f \in \{click, scroll, type\}$. Only the type function needs a value.

Definition 2. \textbf{page content.}
The page content is composed of numerous UI trees \cite{zhang2021screen}. A UI tree is a structured JSON array, encompassing various attributes such as text, position, size, and color. The element of action refers to the text here. It's important to emphasize that the data from UI trees is processed through SecretFlow \cite{ma2023secretflow} for privacy preservation, with only generic content being retained.

Definition 3. \textbf{candidate action elements}
In the process of executing the action prediction, we design a candidate set embedded within the prompt. This particular candidate set is derived from the text corpus of the page content.

\subsection{Object Detection}
Utilizing the raw page content to construct a prompt engenders three problems:
\begin{itemize}
    \item The abundance of the candidate action elements amplifies the difficulty in selecting the correct answer.
    \item Within a page, the potential existence of identical text segments representing distinct meanings compromises the uniqueness of each element.
    \item Original page content consumes a large number of tokens, leading to the overly long context problem.
\end{itemize}

To alleviate the first two problems, we integrate a visual model to comprehend the page content. The information from UI trees \cite{zhang2021screen} allows for a rough reconstruction of the overall page layout, thereby leveraging visual capabilities to analyze the relationships between different action elements. As shown in Figure 2, the date and the price on the calendar form a whole and together point to one meaning, i.e., selecting the flight date. Therefore, when building the candidate action elements, only one needs to be selected. So we optimize the grouping of page content through a bounding box detection model. Moreover, through effective grouping, the surrounding text \cite{gao2005web} can endow candidate action elements with uniqueness. Specifically, we employ the YOLOX \cite{ge2021yolox} model, which is an efficient and practical object detection model that maintains high accuracy. On our dataset, it achieved an excellent performance of \textbf{mAP0.92@IoU75} \cite{ge2021yolox}, demonstrating its powerful performance in practical applications.

To mitigate the third problem mentioned above, we designed a text extraction module that extracts text from the page content, saving a substantial number of tokens.

\subsection{Previous Action Description Generator}
\begin{figure}[h!]
    \centering
    \includegraphics[width=0.5\textwidth]{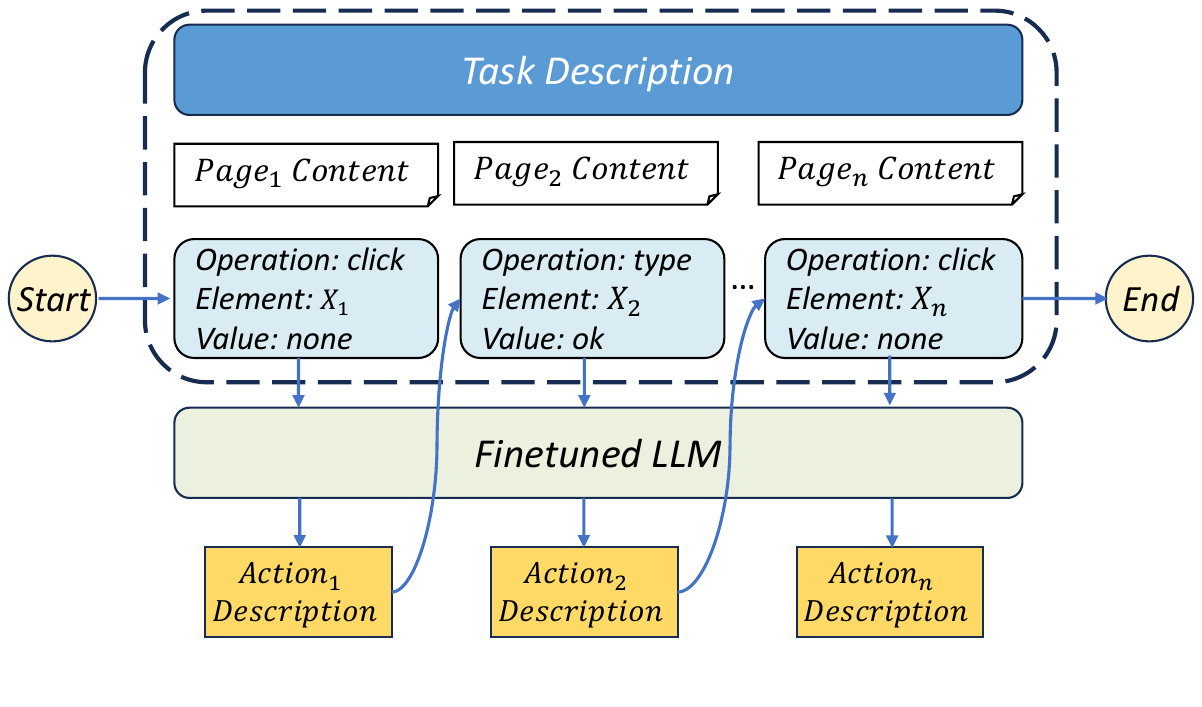}
    \caption{Structure of Previous Action Description Generator.}
    \label{fig:action description}
\end{figure}

Through the Object Detection module, LLMPA Agent is equipped to proficiently delineate the hierarchical structure of the page content. Subsequently, we will introduce another significant feature of the agent: the action-understanding capability. This feature offers two advantages:
\begin{itemize}
    \item It defines the progress of task execution clearly.
    \item It bolsters the context sensitivity, implying that the action prediction model can more proficiently emphasize the influence of earlier operations on succeeding ones.
\end{itemize}
Following prior works \cite{furuta2023multimodal}, in our scenario, the automatic execution of pages can also be described as a deterministic sequential decision problem. Assuming that the present prediction corresponds to the i-th step. We define the fixed instruction as $I$, the prior action space as $\mathcal{A}=\{a_1,a_2,...a_i-1\}$ and the page content space as $\mathcal{P}=\{p_1,p_2,...,p_i\}$. The final function to be optimized is $g(I, A, P)$. For a specific $a_j$ in space $\mathcal{A}$, such as clicking ``ok'', it may signify confirmation of items in a shopping cart, acceptance of a special deal, and so on. Essentially, it is a low-semantic command that lacks context, thereby failing to accurately convey its inherent meaning.

Figure 3 shows the structure of the Previous Action Description Generator. Based on the high-level task description and the corresponding historical behavior sequences on Alipay, the executable key paths can be excavated and represented as action sequences, which are defined as $\mathcal{C} = \{a_1, a_2,..., a_n\}$. $\mathcal{P}$ is the page content space, each $a_i$ corresponds to $p_i$. We adopt a recursive architecture, signifying that comprehending prior actions serves as the input for interpreting the next action. This design makes the semantic expression between continuous actions more coherent. So the function expression for the i-th action is:
\begin{equation}
f_i = z(a_i, p_i, p_{i+1}, f_{i-1}),
\end{equation}
where $f_i$ means the description of i-th action, and z is the LLM-based model. The final output offers insight into both the immediate action behavior and the subsequent page alterations.

\subsection{Instruction Chains Generator}
Drawing from human cognition, we typically decompose intricate tasks~\cite{wei2022chain,yao2023tree} into a series of simpler sub-tasks, allowing for a sequential execution. Inspired by this, we propose the concept of instruction chains, which generates corresponding steps according to the task description. In fact, it provides a comprehensive summary, helping the action prediction model to understand the overall process of the task more clearly and to predict the next action better.

Two kinds of instruction chains are defined:
\begin{itemize}
    \item \textbf{Abstract Instruction Chains.}
    It provides a macroscopic perspective, which helps to simplify complex tasks. This allows the model to comprehend the task process and objectives from a comprehensive view. Next, we will illustrate with a concrete example:\\\\
    \fbox{%
    \begin{minipage}{\linewidth}
    Task: Book an economy class flight ticket from Hangzhou to Beijing on November 4th \\\\
    1. Open a reliable travel booking platform.\\
    2. Enter ``Hangzhou'' as departure, ``Beijing'' as destination, and select November 4th.\\
    3. Choose ``Economy Class'' and search for flights.\\
    4. Select your preferred flight and proceed to book.\\
    5. Fill in the necessary traveler details and go to payment.\\
    6. Complete the payment and receive a confirmation email.
    \end{minipage}%
    }\\
    
    \item \textbf{Elaborate Instruction Chains.}
    It offers detailed guidance, minimizing confusion and ambiguity. As a result, the model is able to predict subsequent actions more easily. Similarly, we will demonstrate this with an example:\\\\
    \fbox{%
    \begin{minipage}{\linewidth}
    Task: Book an economy class flight ticket from Hangzhou to Beijing on November 4th \\\\
    1. Open the Alipay app and log in.\\
    2. Click ``Transport'' on the homepage.\\ 
    3. Input ``Hangzhou'' as the departure city.\\
    4. Input ``Beijing'' as the destination.\\
    5. Set the date to November 4th.\\
    6. Select the ``Economy'' class flight.\\
    7. Search flight and choose an appropriate flight.
    \end{minipage}%
    }
    
\end{itemize}
The most significant distinction between the two examples above lies in the latter's tighter integration with the action space. Naturally, the latter also demands a greater quantity of high-quality samples in the relevant field. To ensure precise execution of tasks, we trained the elaborate instruction chains generation model in our scene. Just like the previous action description generation model, this model is also based on LLM. In contrast, we utilize the abstract instruction chains in the subsequently introduced Agentbench-WB ~\cite{liu2023agentbench} dataset.

\begin{figure}[h!]
    \centering
    \includegraphics[width=0.45\textwidth]{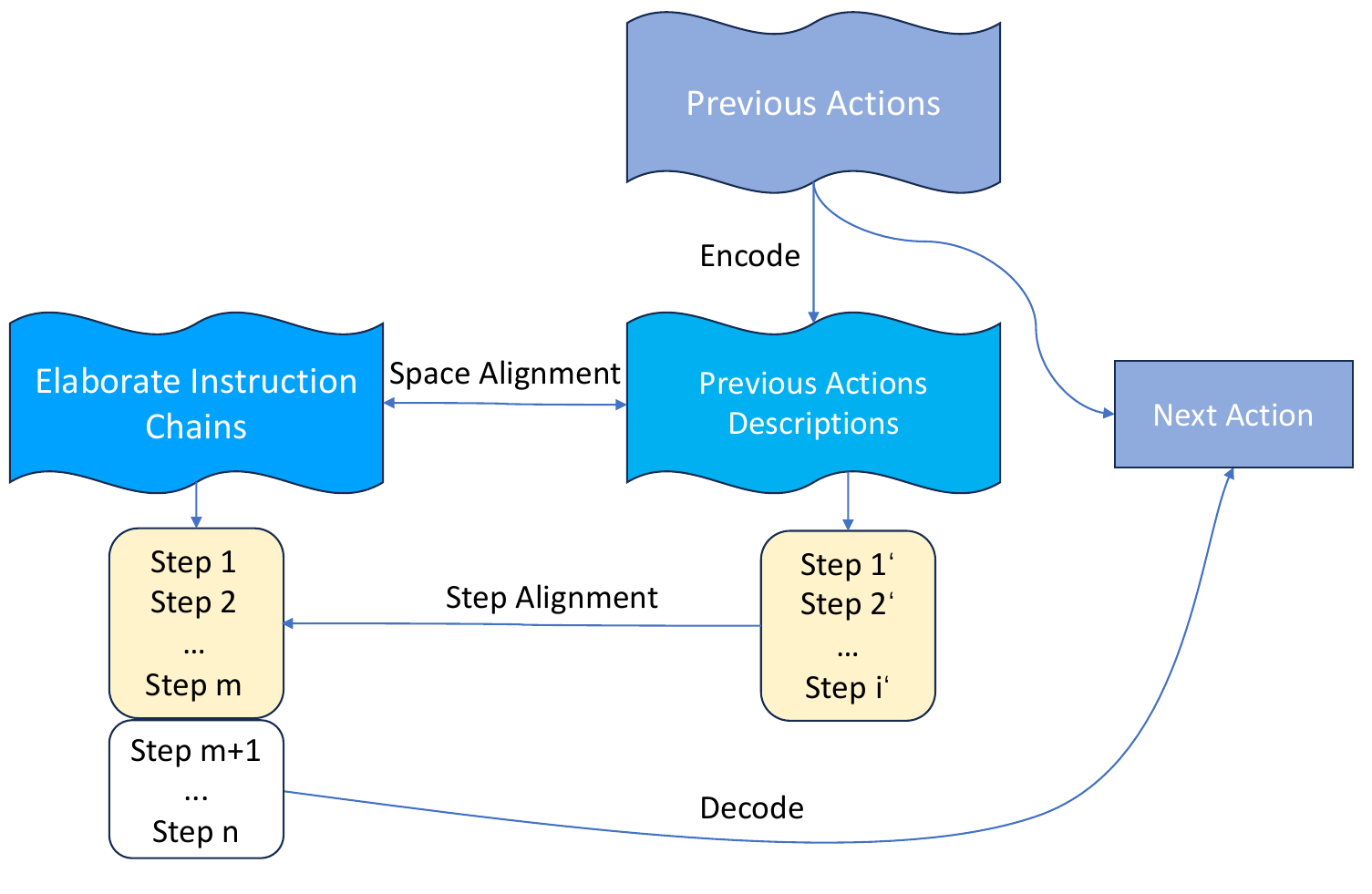}
    \caption{Advantages of Combining Action Descriptions and Elaborate Instruction Chains.}
    \label{fig:analysis}
\end{figure}

Upon analysis, the combination of Previous Action Description(PAD) and Elaborate Instruction Chains(EIC) can significantly reduce the difficulty of the next action prediction. As shown in Figure 4, different colors represent different spaces, and the closer the colors are, the higher the spatial similarity. PAD can be characterized as the encoding process from action to action description. Since the generation of EIC highly depends on the action, it can be approximated that EIC and PAD are spatially similar. Assuming a task is given and has been executed for i-th step. We define the space of EIC as $\mathcal{S}=\{Step_1, Step_2, ...,Step_n\}$, and the space of PAD as $\mathcal{S'}=\{Step_1', Step_2', ..., Step_i'\}$. By calculating semantic similarity, $S'$ can be aligned to a continuous subspace of $\mathcal{S}$, assuming this space is $\mathcal{S}_{\alpha}=\{Step_1, Step_2,...,Step_m\}$, then we have $\mathcal{S'} \simeq \mathcal{S}_{\alpha}$. Therefore, the prediction space for the next step will be narrowed down to $\mathcal{S}_{\beta}=\{Step_{m+1},..., Step_n\}$. This yields explicit assistance for subsequent action prediction.

\subsection{Action Prediction}
Our prompt is an amalgamation of several components: task description, instruction chains, history actions, descriptions of history actions, extracted text, and candidate action elements. The candidate action elements are derived from the top elements of historical data, the task itself, and the extracted text from bounding boxes. We then employ the LLM to train our next-action prediction model.

\subsection{Controllable Calibration}
Due to the hallucination problem of large language models(LLMs), even given a candidate, it may still generate other answers. We designed the controllable calibration module, which is composed of two parts:
\begin{itemize}
    \item \textbf{Is it executable?} A CTR model~\cite{juan2016field} can be trained to verify whether the predicted element can be executed. The element can be treated as an item. If a click or type occurs, it is labeled as a positive sample. This transforms the problem into a classic binary classification problem~\cite{kumari2017machine}. By setting a threshold, we can determine whether the element is executable.

    \item \textbf{Does it make logical sense?} Based on the executable key paths mentioned in the Previous Action Description Generator, we add a constraint. If there is no loop in the original path, it is logically unreasonable for a loop to appear in the result.
\end{itemize}
If the predicted element passes both checks successfully, it will be handed over to the executor for simulation-based operations. However, in case of failure, it will be rerouted back to the Action Prediction module for further output revision.

\section{Experiment}
To evaluate the effectiveness of the LLMPA Agent, we conduct experiments on both real-world online environments within Alipay and public benchmarks.

\subsection{Online Deployment}
For online evaluation, we validate the effectiveness of our framework with Alipay.

% For online evaluation, we evaluate our framework with Alipay.which comprising numbers of services, allowing our framework to engage directly with real-world scenarios. 

\paragraph{\textbf{Environment.}}
Alipay, a prominent global payment tool, encompasses numerous services that cater to various digital life and finance scenarios. Our proposed LLMPA Agent has significantly enhanced the user experience. We have covered more than 20 real-world scenarios in online deployment and manually annotated 2000 different tasks. Extensive online experimentation has effectively demonstrated the superiority of our method.

\paragraph{\textbf{Setting.}}
It should be noted that AntLLM is a pretrained large language model in our work. We report the performance of AntLLM that serves as a baseline 
and LLMPA Agent that integrates components including \textit{Object Detection}, \textit{Instruction Chains \& Previous Action Description}, and \textit{Controllable Calibration}. Regarding \textit{Instruction Chains}, Elaborate Instruction Chains are leveraged considering that in the online environment, we have access to rich and detailed resources of action sequence. We also conducted an ablation study to evaluate the impact of different components on the performance of the LLMPA Agent. The models above are fine-tuned during online experiments.

\paragraph{\textbf{Metric.}} 
We report Success Rate as the evaluation metric including \textit{Step Success Rate (Step SR)} and \textit{Task Success Rate (Task SR)}. A step is considered successful when the selected element and the predicted action correspond to the ground truth. A task is considered successful when all of its constituent steps have succeeded. 

\paragraph{\textbf{Deployment details.}}
Our method was trained using 32 NVIDIA A-100 GPUs. For online inference, we employed 100 NVIDIA A-40 GPUs in the production environment. We conducted weekly finetuning using the latest data to ensure the freshness of model performance.

\begin{table}[!h]
\centering
\caption{Online experimental results with Alipay environment. \textit{Step SR} and \textit{Task SR} stands for \textit{Step Success Rate} and \textit{Task Success Rate}, while \textit{IC \& PAD} stands for \textit{Instruction Chains \& Previous Action Description.} \textit{Baseline} means fine-tuned AntLLM.}
\begin{tabular}{lcc}
\toprule
Method & Step SR & Task SR \\
\midrule
Baseline & 52.42 & 6.05    \\
LLMPA w/o Object Detection & 86.02 & 48.46   \\
LLMPA w/o IC \& PAD & 65.43 & 14.27    \\
LLMPA w/o Controllable Calibration  & 84.53 & 45.85   \\
LLMPA & \textbf{93.71} & \textbf{70.42}   \\
\bottomrule
\end{tabular}
\label{tab:online}
\end{table}

\paragraph{\textbf{Online Performance and Analysis.}}
As shown in Table~\ref{tab:online}, the AntLLM without any additional design obtains the poorest performance as expected. LLMPA outperforms the baseline by a significant margin and achieves a high success rate in terms of step success rate (93.71\%), demonstrating the effectiveness of the framework. Furthermore, the absence of each component leads to a noticeable decrease in performance, with \textit{Instruction Chains \& Previous Action Description} having the most significant impact. This indicates the importance of task decomposition and high-level summarization capabilities.

\subsection{Benchmark Test}

\begin{table}[!t]
\centering
\caption{Experimental results on public benchmark AgentBench-WB. \textit{Step SR}, \textit{Ele. Acc}, and \textit{Op. F1} stands for \textit{Step Success Rate}, \textit{Element Accuracy} and \textit{Operation F1}. \textit{IC} stands for \textit{Instruction Chains}.}
\begin{tabular}{l|c|ccc}
\toprule
Models & IC & Step SR  & Ele. Acc&Op. F1\\
\midrule
\multirow{2}*{\texttt{llama2-7b-chat~\cite{touvron2023llama2} }  } & \XSolidBrush    & 7.25  & 9.38 & 44.71\\
 & \Checkmark    & 8.10  & 9.55 & 45.76\\
\midrule
\multirow{2}*{\texttt{llama2-13b-chat~\cite{touvron2023llama2}}} & \XSolidBrush    & 9.29  & 10.40 & 47.36\\
 & \Checkmark    & 10.27  & 11.96 & 47.52\\
 \midrule
\multirow{2}*{\texttt{baichuan2-7b-chat~\cite{baichuan2023baichuan2}}} & \XSolidBrush    & 10.49  & 10.91 & 48.22\\
 & \Checkmark    & 11.51  & 13.81 & 46.79\\
 \midrule
\multirow{2}*{\texttt{gpt-3.5-turbo~\cite{openai2022chatgpt}}} & \XSolidBrush    & 16.79  & 21.23 & 47.45\\
 & \Checkmark    & 17.82  & 22.51 & 45.09\\
  \midrule
 \multirow{2}*{\texttt{gpt-4~\cite{openai2023gpt4}}} & \XSolidBrush    & 24.65  & 33.47 & 52.94\\
 & \Checkmark    & 26.36  & 35.23 & 52.65\\
\bottomrule
\end{tabular}
\label{tab:offline}
\end{table}

We conduct experiments on public datasets to evaluate the effectiveness of the LLMPA.
\paragraph{\textbf{Environment.}}
We select the Web Browsing environment from AgentBench (namely AgentBench-WB)~\cite{liu2023agentbench} as the representative benchmark. AgentBench-WB, derived from Mind2Web dataset~\cite{deng2023mind2web}, encompasses 912 tasks from 73 websites spanning domains such as Housing, Job, Social Media, Education, etc. 

\paragraph{\textbf{Setting.}}
Evaluation on AgentBench-WB\cite{liu2023agentbench} involves two stages: ranking HTML elements with a fine-tuned small language model and predicting action in the form of multi-choice QA with an agent. The ranking models employed in our experiments align with AgentBench, and the primary focus of offline experiments is to evaluate the capability of action prediction. It is worth noting that there still remains a gap between the simulation environment and real-world scenarios in this area. For instance, in comparison to the online environment, AgentBench-WB offers a relatively limited number of web pages and lacks dynamic interaction capability. Consequently,  it is not feasible to implement all of the methods we proposed in the simulation environment. Therefore, we conduct the evaluation solely focusing on the effectiveness of \textit{Instruction Chains}, which is proved to be a core component of LLMPA. We employ \textit{Instruction Chains} in the form of Abstract Instruction Chains, for the limited resource of action sequences in a simulation environment. Evaluations are conducted in the manner of in-context learning. 

\paragraph{\textbf{Metric.}}
\textit{Step Success Rate} is employed as an evaluation metric. Following Mind2Web\cite{deng2023mind2web}, we also report \textit{Element Accuracy} that calculates the accuracy of the selected element, and \textit{Operation F1} that calculates the token-level F1 score for the predicted operation. 

\paragraph{\textbf{Implementation details.}}
All experiments of benchmark testing run on the Linux server(Ubuntu 16.04) with the Intel(R) Xeon(R) Silver 4214 2.20GHz CPU, 512GB memory, and 8 NVIDIA A-100 GPUs.

\paragraph{\textbf{Results.}}
From the results presented in Table~\ref{tab:offline}, we observe that the leverage of \textit{Instruction Chains} contributes to enhancing performance on both open-sourced and API-based models, suggesting the effectiveness and generalization capability of the method across different models. Meanwhile, in comparison to online experiments, the improvements brought by \textit{Instruction Chains} are less pronounced, indicating that more specific instruction chains lead to higher performance.

\subsection{Case Study}
\begin{figure}[t!]
    \centering
    \includegraphics[width=0.45\textwidth]{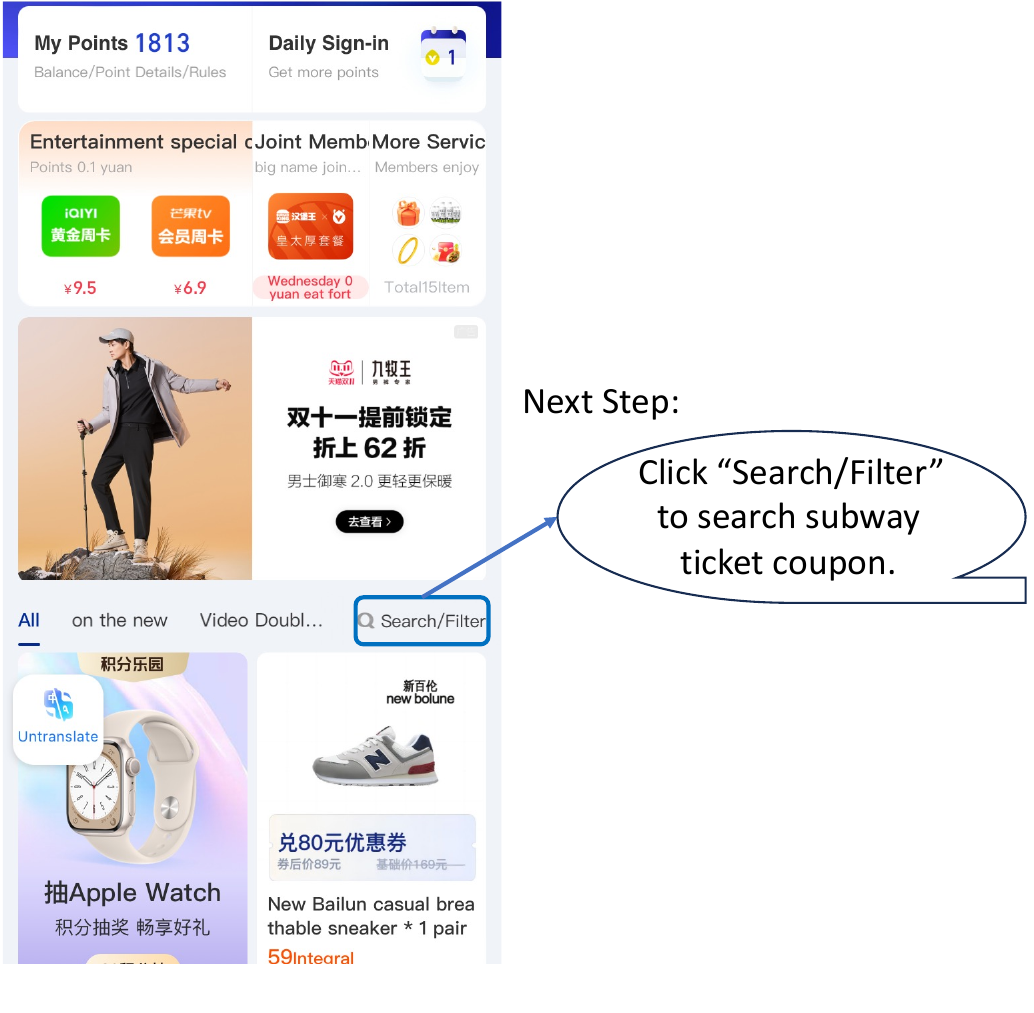}
    \vspace{-4mm}
    \caption{Step to search subway ticket coupon.}
    \vspace{-4mm}
    \label{fig:step_search}
\end{figure}

To validate the effectiveness of our proposed methodology, we undertake an analysis of a practical application, specifically, the process of redeeming subway discount vouchers in the Alipay membership scenario. Figure 5 illustrates one step of the full process where normally expect to click the search bar and type subway ticket coupon. It is significantly difficult for the model to infer the underlying significance of this search bar by relying solely on the contextual information, leading to incorrect predictions. From a certain perspective, instruction chains could be regarded as an invaluable enhancement of external knowledge, which offers pivotal guidance for the action prediction model to comprehend the task's inherent operational logic.

\begin{figure}[h!]
    \centering
    \includegraphics[width=0.45\textwidth]{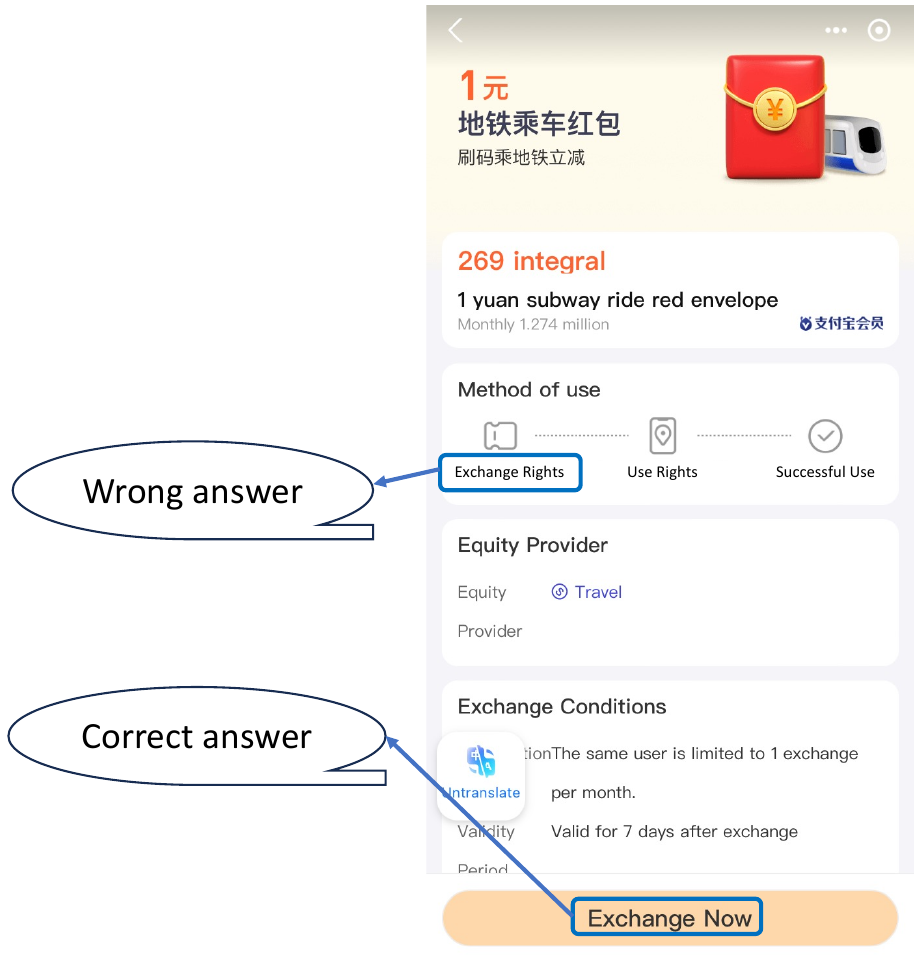}
    \vspace{-4mm}
    \caption{Step to exchange subway ticket coupon.}
    \vspace{-4mm}
    \label{fig:step_exchange}
\end{figure}
As depicted in Figure 6, the accurate action entails clicking on ``Exchange Now''. However, the action prediction model erroneously returns ``Exchange Rights''. Even though these two elements bear a high degree of semantic similarity, they differ significantly in terms of operability. In such instances, the controllable calibration module is adept at discerning that ``Exchange Rights'' is non-clickable and non-typable. Upon receiving this feedback, the model refines its output accordingly. Through this mechanism, the model can effectively sidestep a host of glaring errors.

\section{Discussion}
The proposed virtual assistant system based on large language models shows promise in its ability to parse complex instructions, reason about goals, and execute chained tasks autonomously. However, there are both advantages and disadvantages to this approach that warrant further discussion.

A key advantage is the enhanced natural language processing and reasoning capabilities enabled by the large language model architecture of the LLMPA module. By leveraging large amounts of training data, the LLM is able to comprehend ambiguous or incomplete instructions and better infer user intent. This allows the assistant to successfully interpret and act on a wider range of natural language requests. Additionally, the LLM's ability to break down goals and predict the next actions facilitates multi-step procedural task completion.

However, there are also limitations to relying solely on the LLM. The system remains constrained by the training data, meaning unfamiliar phrasings or requests may still pose a challenge. There are also risks of biased or erroneous behavior if the training data contains imperfections. From an implementation standpoint, large language models can be resource-intensive, making deployment on mobile devices challenging. More work is needed to optimize the models for on-device usage.

Future work should focus on expanding the training data to encompass more diverse user utterances and reducing the resource requirements for mobile deployment. Exploring different model architectures that combine the strengths of LLMs with other techniques may further enhance reasoning abilities. Ongoing improvements to the contextual modules and executor components will also contribute to more intelligent automated assistants. Evaluating the system's capabilities and limitations through real-world testing with end users at scale is critical.

\section{Conclusion}
This work introduced a novel approach to intelligent virtual assistants using large language models designed specifically for mobile app automation. We proposed an end-to-end architecture consisting of the LLMPA model, environmental context, and an executor that enabled automated multi-step task completion in a real-world payment app based on natural language instructions. Testing at a large scale in the widely used Alipay platform demonstrated that modern large language models can power assistants capable of understanding goals, planning, and accomplishing intricate real-world procedures to assist users. This represents a major advance for intelligent assistants and their adoption in ubiquitous mobile applications. Further development of contextual processing, reasoning capabilities, and optimized on-device deployment can build on these foundations to realize the full potential of virtual agents able to comprehend language, plan, and take action to aid humans in their daily lives.

\bibliographystyle{ACM-Reference-Format}
\bibliography{LLMPA}

% \appendix

\end{document}